\documentclass[12pt]{iopart}

\usepackage{iopams}  
\usepackage{appendix}
\usepackage{natbib} 
\usepackage{verbatim}
\usepackage[noadjust]{cite}

\usepackage{graphicx}
\usepackage[colorlinks = true,
            linkcolor = blue,
            urlcolor  = blue,
            citecolor = blue,
            anchorcolor = blue]{hyperref}

\begin{document}

\title{Deep Dose Plugin: Towards Real-time Monte Carlo Dose Calculation Through a Deep Learning–based Denoising Algorithm}

\author{Ti Bai, Biling Wang, Dan Nguyen, Steve Jiang}

\address{Medical Artificial Intelligence and Automation (MAIA) Laboratory\\ Department of Radiation Oncology\\
University of Texas Southwestern Medical Centre, Dallas, Texas 75390}
\ead{Steve.Jiang@UTSouthwestern.edu}
\vspace{10pt}

\begin{abstract}
Monte Carlo (MC) simulation is considered the gold standard method for radiotherapy dose calculation. However, achieving high precision requires a large number of simulation histories, which is time-consuming. The use of computer graphics processing units (GPUs) has greatly accelerated MC simulation and allows dose calculation within a few minutes for a typical radiotherapy treatment plan. However, some clinical applications demand real-time efficiency for MC dose calculation. To tackle this problem, we have developed a real-time, deep learning–based dose denoiser that can be plugged into a current GPU-based MC dose engine to enable real-time MC dose calculation. We used two different acceleration strategies to achieve this goal: 1) we applied voxel unshuffle and voxel shuffle operators to decrease the input and output sizes without any information loss, and 2) we decoupled the 3D volumetric convolution into a 2D axial convolution and a 1D slice convolution. In addition, we used a weakly supervised learning framework to train the network, which greatly reduces the size of the required training dataset and thus enables fast fine-tuning–based adaptation of the trained model to different radiation beams. Experimental results show that the proposed denoiser can run in as little as 39 ms, which is ~11.6 times faster than the baseline model. As a result, the whole MC dose calculation pipeline can be finished within $\sim$0.15 seconds, including both GPU MC dose calculation and deep learning–based denoising, achieving the real-time efficiency needed for some radiotherapy applications, such as online adaptive radiotherapy. 
\end{abstract}

%
%

\section{Introduction}
\label{sec:introduction}
Accurate dose calculation is vitally important for the success of modern radiotherapy techniques, such as intensity-modulated radiation therapy (IMRT) \citep{RN1,RN2} and volumetric modulated arc therapy (VMAT) \citep{RN7,RN8}. Monte Carlo (MC) simulation is considered the most accurate dose calculation algorithm \citep{RN14,RN15}. However, since MC simulation is a stochastic process, the resulting dose map contains inherent quantum noise whose variance is inversely proportional to the number of the simulation histories and, accordingly, to the simulation time. Typically, achieving clinically acceptable precision requires hours of CPU computation time. Graphics processing unit (GPU)–based parallel computation frameworks can accelerate MC simulation to a few minutes for a typical IMRT/VMAT plan \citep{RN23,RN24}. This meets the requirements for a final dose calculation engine in the clinical workflow, where the final plan is recalculated with an accurate dose calculation prior to patient delivery. However, several areas in the clinical workflow require real-time dose calculation, such as inverse optimization of the treatment planning process for IMRT and VMAT. The importance of real-time dose calculation has been amplified by the push for online adaptive radiotherapy, where patients must be completely re-planned within a matter of minutes for a given treatment fraction to account for changing anatomy—such as tumor shrinkage or weight loss—over the course of treatment \citep{RN30,RN31}. Since the dose calculation is only a small fraction of the re-planning process, it is essential to minimize its duration by accelerating GPU-based MC dose calculation from minutes to sub-seconds.

Inspired by its unprecedented successes in various fields \citep{RN44,RN45}, researchers have used convolutional neural network (CNN)–based deep learning (DL) techniques to suppress the quantum noise in MC simulations obtained with a small number of histories \citep{lin2020detail,fornander2019denoising,xu2019adversarial,RN46,RN48,RN49,RN50}. For instance, in the radiation therapy field, Javaid et al. \citep{RN48} proposed a UNet-based \citep{RN46} denoiser for proton therapy. This denoiser is trained by using the MC dose results for $1\times{10}^6$ particles as the input and the MC dose results for $1\times{10}^9$ particles as the target. Peng et al.\citep{RN49} demonstrated a convolutional encoder-decoder neural network’s ability to accelerate the MC radiation transport simulation for patient dose calculation for computed tomography imaging. Neph et al. \citep{RN50} trained a multi-branches network for magnetic resonance guided radiotherapy by simultaneously feeding the noisy MC dose, MC fluence and CT geometry into different branches, where each branch is a UNet.

To our knowledge, the existing DL-based denoisers have yet to reach the real-time efficiency for 3D dose denoising discussed in this work. For example, the UNet-based denoiser proposed by Javaid et al. \citep{RN48} processed the dose volume in a 2.5D way, i.e., by processing multiple ($<$7) slices simultaneously and then processing the next set of multiple slices. They reported an inference time of ~10 s, given an image size of $512\times512$, when using an NVIDIA Titan X GPU. In this paper, we aim to bring the DL model inference efficiency to the real-time level (sub-seconds) without compromising its denoising performance.

In addition, we will address issues related to the application of the developed DL-based MC dose denoiser to different hospitals, treatment machines, and treatment beams. Transfer learning is an effective way to mitigate performance degradation when there is a distribution shift between the source domain and the target domain. One simple way to implement transfer learning is to fine-tune a pre-trained model by using extra data from the target domain. For this approach, end users would need to generate the extra data required to fine-tune the model before they could use the model in their clinical practice. With regard to the MC dose denoising task studied in this work, in the standard supervised learning setting, we would have to generate a massive set of paired noisy and clean dose maps. Generating these data would be time-consuming, especially for the clean MC dose distributions calculated with a large number of histories. This would be burdensome to end users and would thus hamper the clinical implementation of the DL-based MC dose denoiser we developed. Alternatively, Lehtinen et al. \citep{RN51} proposed a noise-to-noise weakly supervised training scheme for natural image processing and showed denoising performance comparable to the conventional noise-to-clean supervised learning setting. Inspired by this work and by the need to make the developed model more clinically practical, we introduce a weakly supervised learning framework in which both the input and the target are noisy dose maps.

Because real-time DL-based dose denoising is needed for certain clinical applications, such as IMRT iterative plan optimization and online adaptive re-planning, we have developed a real-time DL-based dose denoising plugin, termed herein as deep dose plugin. Our deep dose plugin has plug-and-play ability, so it can be easily inserted into a current treatment planning system (TPS) for dose calculation to quickly denoise the noisy MC dose map. We introduce multiple strategies to ensure that our deep dose plugin is fast in both the training and the inference stages. We have devised a new lightweight CNN by using a multichannel input strategy and exploiting a 3D volumetric convolutional kernel decoupling strategy, which facilitates the real-time inference speed of our deep dose plugin. We also use a noise-to-noise weakly supervised learning framework to enable fast fine-tuning–based model adaptation for easy and fast clinical implementation. 

\section{Methods and Materials}
\subsection{Methods}
\subsubsection{Noise-to-noise training scheme}
Let us first formulate our problem mathematically. Given a noisy volumetric dose map, $Y\in R^{H\times W\times D}$, with $H$ rows, $W$ columns and $D$ slices, that is corrupted by quantum noise, $\epsilon\in R^{H\times W\times D}$, due to MC dose calculation with a small number of histories, our task is to restore the underlying clean signal, $X\in R^{H\times W\times D}$, from $Y$, where we have the following relationship:
\begin{equation}
  Y=X+\epsilon.
  \label{eq:noise_equation}
\end{equation}
From the perspective of supervised deep learning, problem~(\ref{eq:noise_equation}) can be solved by training a CNN, $\Phi$, with parameters, $W$, so that the output of $\Phi_W(Y)$ is as close to $X$ as possible. The associated cost function can be expressed as:
\begin{equation}
    W=\arg\min_{W}\int_{X}\int_{\epsilon}||\phi_W(X+\epsilon)-X||_{2}^{2}d\epsilon dX
    \label{eq:DL_objective}
\end{equation}

From cost function~(\ref{eq:DL_objective}), it can be seen that both the clean dose map, $X$, and the associated noise, $\epsilon$, should be provided to train the network. For our volumetric MC dose denoising task, one needs to track a massive number of simulation histories to generate a “clean” dose map. This data preparation process requires a long time, which limits the practical utility of denoising. Luckily, this clean signal requirement can be relaxed to another noisy counterpart of $X$. In other words, the network, $\Phi$, can be trained by using a strategy of noisy input $Y_1=X+\epsilon_1$ to anther noisy target $Y_2=X+\epsilon_2$ instead of the conventional noisy input $Y=X+\epsilon$ to clean target $X$ strategy, where the subscripts 1,2 denote two different MC dose calculations with small numbers of histories. For this noise-to-noise training scheme, the cost function can be expressed as:

\begin{equation}
    W=\arg{\min_W{\int_{X}\int_{\epsilon_1}\int_{\epsilon_2}\left|\left|\Phi_W\left(X+\epsilon_1\right)-\left(X+\epsilon_2\right)\right|\right|_2^2d_{\epsilon_2}d_{\epsilon_1}dX}} 
    \label{eq:DL_objective_N2N}
\end{equation}

This attractive strategy has been used in a previous work\citep{RN51}. We would like to provide a more comprehensive proof, as shown in~\ref{appendix: equivalence}. 

This noise-to-noise weakly supervised learning setting lays the foundation for a fast data generation stage and can thereby reduce the overall training time. This property is important for adapting the model by fine-tuning during clinical implementation.

\subsubsection{Lightweight CNN architecture}
Our deep dose plugin has a similar network architecture to the popular 3D UNet-based architecture. However, the inference time of the conventional 3D UNet architecture cannot achieve the desired real-time efficiency. One may decrease the computational burden by decreasing the input size and/or the model size, but doing so would usually compromise the model performance. First, a smaller input size would mean representing the same field-of-view with a larger voxel size. This would inevitably lead to the loss of fine details that are only visible in the fine scale representation. Consequently, the model could not utilize these fine details for prediction, which would lead to inferior model performance. On the other hand, a lighter model usually suggests higher inference efficiency. One typical technical line towards a lighter model is to reduce the model depth. For instance, one can reduce the deep residual network (ResNet) \citep{RN44} from 152 layers to 18 layers and thereby substantially increase the inference speed given the same input size. Nonetheless, reducing the model depth would also mean reducing the receptive field, which would yield inferior denoising ability. The reason is that richer sematic information, and hence a larger receptive field, is required to capture the long-range correlations among voxels, which is of vital importance for denoising. Moreover, reducing the model depth would also hamper the nonlinear expressiveness of the network because of the smaller number of nonlinear layers. Thus, one cannot simply decrease the input size or reduce the model depth to accelerate the deep learning model.

\begin{figure}
    \centering
    \includegraphics[width=\textwidth]{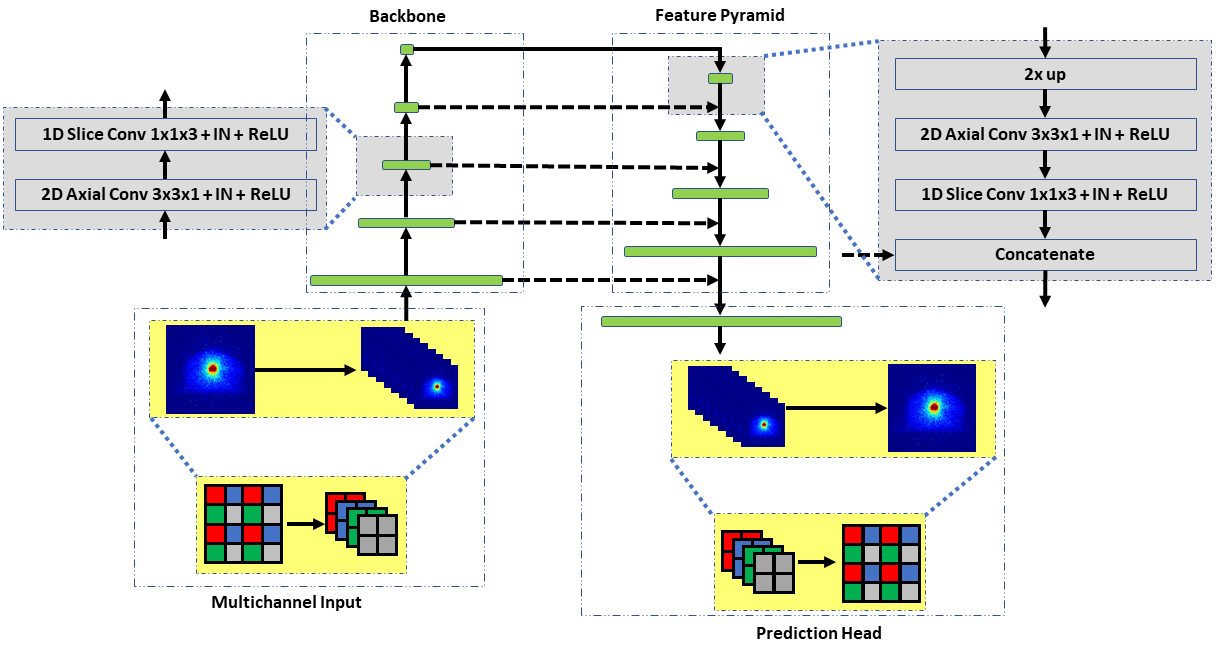}
    \caption{Network architecture illustration. There are four major components: multichannel input, backbone, feature pyramid and prediction head. The multichannel input is consisting of a voxel unshuffle operator that can effectively reduce the feature map size. In the backbone and feature pyramid, all the 3D volumetric convolution operators are decomposed into a 2D axial convolution operator and a 1D slice convolution operator. The prediction head contains another decomposed convolutional layer and a voxel shuffle operator to restore the final one-channel dose map.}
    \label{fig:network}
\end{figure}
To reach a real-time deep dose plugin, we introduce several strategies in this paper to accelerate the conventional UNet-based architecture. As illustrated in Figure~\ref{fig:network}, we first use a voxel unshuffle layer to rearrange the original input with a dimension of $B\times1\times H \times W \times D$ into an 8-channel input with a dimension of  $B\times8\times\frac{H}{2}\times\frac{W}{2}\times\frac{D}{2}$. This can substantially reduce the feature map sizes of all the subsequent layers, and thereby reduce the computation burden. Then, for all the convolutional operators, we decompose the regular 3D volumetric convolution operator into a 2D axial convolution operator and a 1D slice convolution operator. This can further decrease the computation complexity. Lastly, we adopt a 3D voxel shuffle operator to restore the final one-channel dose map output, which is the inverse operator of the voxel unshuffle operator. More details can be found from~\ref{appendix: architecture}.

\subsection{Materials}
\subsubsection{Dataset description}
A dataset consisting of 199 prostate patients with IMRT plans was used to generate volumetric MC dose maps for algorithm verification. This data pool was randomly split into 162 cases as the training dataset and 37 cases as the validation dataset. For training purposes, two independent noisy MC simulations, each with $1\times{10}^6$ histories, were conducted for each patient case in the training dataset. For validation purposes, we conducted 15 independent noisy MC simulations, each with $1\times{10}^6$ photon histories, for each patient case in the validation dataset. We used multiple noisy MC simulations of each patient to show the stability of the proposed denoiser. In addition, for quantitative comparison, we also performed one “clean” simulation with $1\times{10}^9$ histories for each patient, which served as the ground truth dose map.

\subsubsection{Training details}
To stabilize the training process, all patient doses were normalized by dividing by 80 Gy, which was the average prescription dose for the patients in the training dataset. All dose maps were resized to a voxel size of $2.34\times2.34\times3.00\ \mathrm{mm}^3$, then fed into the network for training. Then, a $256\times256\times64$ volume containing the effective region was cropped out from the original MC dose map. To augment the training dataset, this cropped volume was padded with 16 zeros outside each dimension, then a random cropping strategy was applied to sample a $256\times256\ \times64$ volume for training.

To train our network via the noise-to-noise weakly supervised learning framework, the input and the target were set as the two noisy simulations, each with one million histories. It should be noted that during the training process, the input and the target were randomly exchanged with each other between these two noisy simulations. This approach can further augment the training dataset size, and its feasibility is theoretically guaranteed by the noise-to-noise weakly supervised learning framework.

Adam optimizer\citep{Rn58} with an initial learning rate of $1\times{10}^{-4}$ was employed to train the network by $2\times{10}^5$ iterations on one Tesla K80 GPU. The learning rate was reduced by 10 times at iterations $1\times{10}^5$ and $1.5\times{10}^5$, corresponding to $1\times10^{-5}$ and $1\times{10}^{-6}$, respectively. The batch size was 1. The hyperparameters associated with the Adam optimizer were $\beta_1=0.9$ and $\beta_2=0.999$.

\subsubsection{Conventional UNet-based denoiser}
To benchmark the performance of the proposed architecture, we also trained a UNet-based MC dose denoiser. Following the default settings, we designed the UNet as the 6-downsampling version since the smallest dimension of the data is 64. Each layer consisted of three operators: regular 3D convolution, instance normalization and ReLU. The kernel size was $3\times3\times3$. The initial feature number was 64. More model configuration parameters can be found in Table~\ref{tab:layer_configuration}.
All other training details for the UNet-based dose denoiser, such as data preprocessing, data augmentation, weakly supervised learning setting, optimizer and associated hyperparameter settings, were the same as for the proposed denoiser. 

\subsubsection{Model performance comparison}
We first qualitatively compared the performance by showing the dose difference maps between the denoised and the “clean” MC dose maps and plotting the representative 1D profiles. Then, we used the mean squared errors (MSE) to quantify the difference between a dose map $X^{\mathrm{denoised}}$ and the benchmarked dose map\ $X^{\mathrm{GT}}$:
\begin{equation}
    \mathrm{MSE}=||X^{\mathrm{denoised}}-X^{\mathrm{GT}}||_{2}^{2}
\end{equation}

In the clinic, dose-volume histograms (DVH) are routinely used for radiotherapy treatment planning. As suggested in the literature \citep{RN59}, one can use the absolute difference in area between two DVHs associated with $X^{\mathrm{denoised}}$ and $X^{\mathrm{GT}}$ to quantify the denoising performance:
\begin{equation}
    \mathrm{DVH\ Error}\ =\ \sum\left|h_i^{\mathrm{denoised}}-h_i^{\mathrm{GT}}\right|\Delta Di
\end{equation}
where $\Delta Di$ is the width of $i_{\mathrm{th}}$ DVH bin and $h_i^{\mathrm{{denoised,\ GT}}}$ represents the values of the two histograms in that bin.

We also used some other clinically relevant indicators to show our algorithm’s performance. We first defined D\# as the minimal dose that is received by \#\% of the planning target volume (PTV). The number \# is defined as a percentage of the prescription dose. We compared PTV D95, PTV D98 and PTV D99 as clinical evaluation criteria. Their relative differences were also reported as $|\frac{\mathrm{D\#_{denoised}-D\#_{GT}}}{\mathrm{D\#_{GT}}}|$, where $\mathrm{D\#_{denoised}}$ and $\mathrm{D\#_{GT}}$ corresponds to the denoised and the ground truth dose volumes, respectively. Gamma passing rate \citep{wendling2007fast} is another widely used metric in clinic. In this paper, we used the $2\%$/2mm criterion to calculate the gamma passing rate for each dose volumes when comparing with the ground truth dose volumes.

We also compared the isodose coverage similarity, which is one of the important metrics that indicates two plans’ similarity. Isodose volume is defined as a binary mask whose value is 1 if the voxel contains a dose value above a specified threshold and 0 otherwise. Let us denote $\mathrm{V}_{\mathrm{\#\%ISO}}$ as the volume of the \#\% isodose region. Then, the isodose coverage similarity with a dose level of \#\% between two dose maps $X^{\mathrm{denoised}}$ and $X^{\mathrm{GT}}$ can be defined as the Dice coefficients between the associated isodose volumes:
\begin{equation}
    \mathrm{Isodose\ similarity}=\ 2\ast\frac{\mathrm{V}_{\mathrm{\#\%ISO}}^{\mathrm{denoised}}\cap\mathrm{V}_{\mathrm{\#\%ISO}}^{\mathrm{GT}}}{\mathrm{V}_{\mathrm{\#\%ISO}}^{\mathrm{denoised}}+\mathrm{V}_{\mathrm{\#\%ISO}}^{\mathrm{GT}}}
\end{equation}
For a more comprehensive comparison, we used six different dose levels (10\%, 30\%, 50\%, 70\%, 80\% and 90\%) to calculate the average isodose similarities.

It should be noted that, to remove the effect of different prescription doses for different patients, we first normalized the dose maps by dividing by the PTV D95. Then, we calculated different metrics against the ground truth dose map, including MSE, DVH error, D95, D98, D99 and isodose similarity for all 15 independent MC simulations of each patient. Patient-specific boxplots for each metric were depicted to compare the performance.

Finally, for the purpose of an overall comparison, we calculated the mean values and the standard deviations (STDs) for the average value of each metric among all the 37 patients, based on the 15 different MC simulations.  

\subsubsection{Computational efficiency comparison}
We compared the computational complexity with both a theoretical value and a time measurement. Specifically, we used the number of float operations (FLOPs) involved for data flowing through a network to theoretically quantify the computation required of any specific network. The FLOPs can be defined as the number of multiplications and summations. For instance, given a convolution operator with a kernel size of $\mathrm{k_h\times k_w\times k_d}$, and given an input of size $\mathrm{1}\times\mathrm{C}_{\mathrm{in}}\times\mathrm{H}\times\mathrm{W}\times\mathrm{D}$, if the output size is $1\times\mathrm{C}_{\mathrm{out}}\times\mathrm{H}_{\mathrm{out}}\times\mathrm{W}_{\mathrm{out}}\times\mathrm{D}_{\mathrm{out}}$, then the FLOPs can be calculated as:
\begin{equation}
    \mathrm{FLOPs=2\times C_{in} \times k_h \times k_w \times k_d \times C_{out} \times H_{out} \times W_{out} \times D_{out}}
\end{equation}
For a further comparison of model complexity, we also provide the number of trainable parameters for both the proposed architecture and the conventional UNet-based architecture (Table~\ref{tab:metrics}). 
In addition, we report the time required for an MC dose simulation and the inference time required to conduct a complete forward propagation of our proposed denoiser and the baseline UNet-based denoiser (Table~\ref{tab:metrics}). All the data are obtained based on one Tesla K80 GPU and calculated from 100 repeated measurements.

\section{Results}

\begin{figure}
    \centering
    \includegraphics[width=\textwidth]{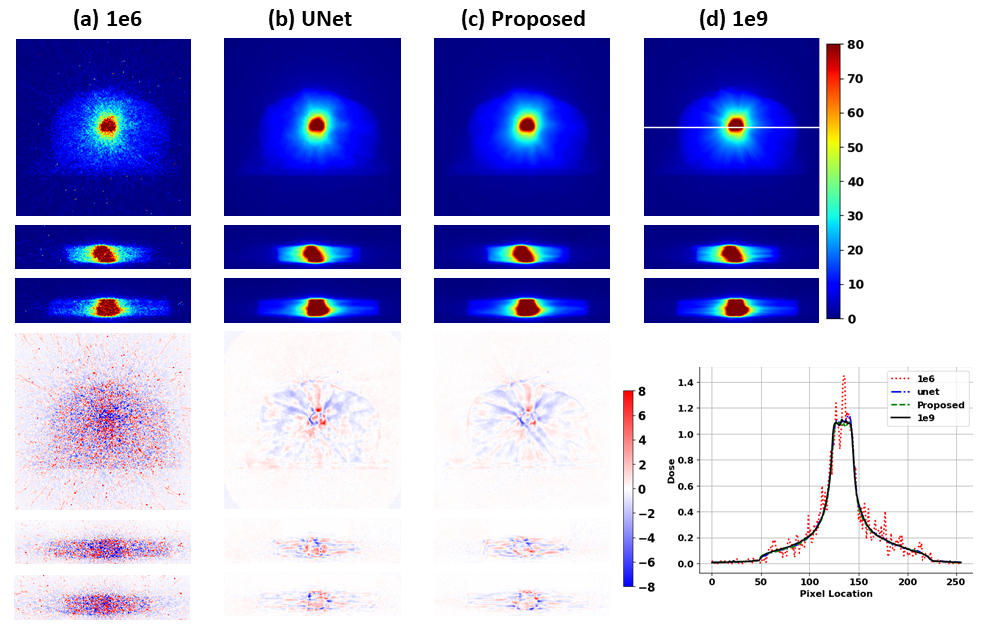}
    \caption{Denoising performance comparison. Dose maps corresponding to (a) the $1\times{10}^6$ photons-based MC simulation, (b) the UNet-based denoiser, (c) the proposed denoiser, and (d) the $1\times{10}^9$ photons-based MC simulation. The display window is [0 80] Gy. The corresponding dose difference maps against the dose maps associated with the $1\times{10}^9$ photons-based MC simulation are presented in the bottom three rows, with a display window of [-8 8] Gy. For both the original and the dose difference maps, the axial, coronal and sagittal views are displayed from the top down. The horizontal profiles across dose difference maps are plotted in the lower right. The location of the profile is indicated by the solid line in (d).}
    \label{fig:diff_profile}
\end{figure}

We first illustrate the denoising performance obtained by the different methods. Figure~\ref{fig:diff_profile}(a) shows strong quantum noise in the noisy dose maps associated with the $1\times{10}^6$ photons-based MC simulation. This statistical fluctuation can be greatly reduced by using substantially more simulation histories, as in the case depicted in Figure~\ref{fig:diff_profile}(d) with $1\times{10}^9$ photons. On the other hand, both the UNet-based denoiser and the proposed denoiser effectively suppressed the noise in the $1\times{10}^6$ photons-based dose map to produce dose maps that are visually comparable to the $1\times{10}^9$ photons-based dose maps. Their denoising performance can be further verified by inspecting their difference maps. The difference maps corresponding to the low photon-based MC simulation exhibit strong noise, while the difference maps with respect to two different denoisers show much fewer differences. To provide a more detailed qualitative comparison, we plotted the horizontal profiles across different dose maps. We can see that both denoisers produced results that match well with the “clean” MC simulation in the low-dose region. In the high-dose region, the proposed method slightly outperformed the UNet-based denoiser, though both denoisers show larger differences in the high-dose region than in the low-dose region. Upon carefully inspecting the profiles and the difference maps further, we find that these dose differences are mainly focused in the large dose gradient regions where denoising is expected to be harder. Overall, in these large dose gradient regions, the proposed denoiser demonstrates slightly better performance than the UNet-based denoiser.

\begin{figure}
    \centering
    \includegraphics[width=\textwidth]{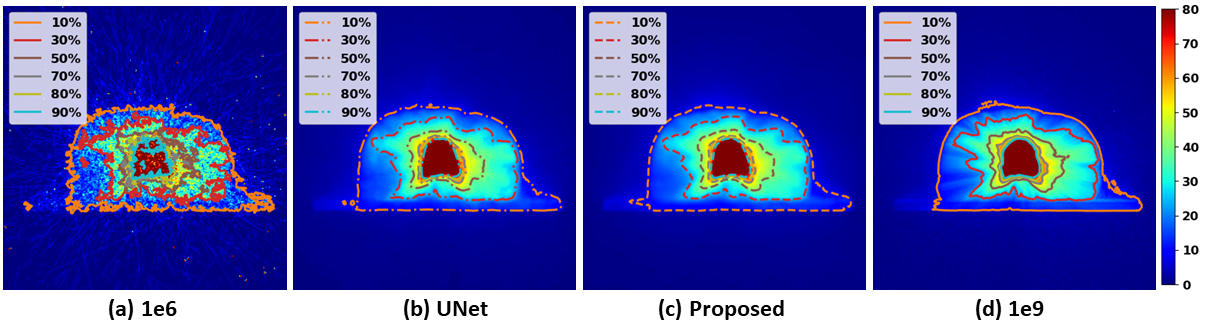}
    \caption{Comparison of isodose line coverage. Dose maps corresponding to (a) the $1\times{10}^6$ photons-based MC simulation, (b) the UNet-based denoiser, (c) the proposed denoiser, and (d) the $1\times{10}^9$ photons-based MC simulation. The display window is [0 80] Gy. Six isodose levels, 10\%, 30\%, 50\%, 70\%, 80\% and 90\%, are plotted and overlaid on the origin dose maps.}
    \label{fig:isodose_line}
\end{figure}
	
To further demonstrate the denoising performance, we present the isodose lines overlaid on the associated dose maps (Figure~\ref{fig:isodose_line}). It should be noted that, for better visualization, we show only the five largest isodose regions of each isodose level for the $1\times{10}^6$ photon-based MC simulation. Both the UNet-based denoiser and the proposed denoiser produced isodose lines comparable to the “clean” MC simulation, especially in the high-dose regions, which are the most clinically relevant.

\begin{figure}
    \centering
    \includegraphics[width=\textwidth]{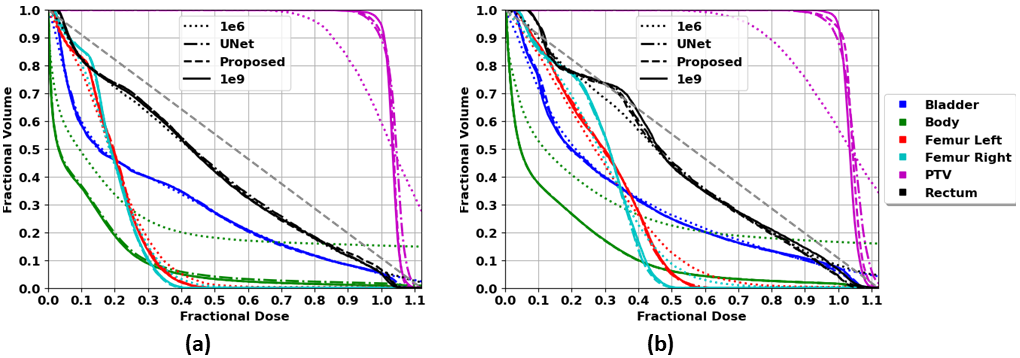}
    \caption{Comparison of DVH profiles associated with (a) the patient case in Figure 2 and (b) the patient case in Figure~\ref{fig:isodose_line}: the $1\times{10}^6$ photons-based MC simulation (dotted line), the UNet-based denoiser (dot-dash line), the proposed denoiser (dashed line) and the $1\times{10}^9$ photons-based MC simulation (solid line). The gray dashed line is the antidiagonal line for reference.}
    \label{fig:dvh}
\end{figure}

Figure~\ref{fig:dvh} plots the DVHs corresponding to the two patient cases presented in Figures~\ref{fig:diff_profile} and~\ref{fig:isodose_line}. As expected, we observed two properties of the histogram with respect to the noisy dose map: 1) many regions (voxels) have a dose larger than 1.2, as indicated by the truncated dotted lines; and 2) the histograms associated with the noisy dose maps are more likely to be antidiagonal (as shown by the gray line in Figure~\ref{fig:dvh}) due to the nature of the quantum noise, which would “uniformly” spread out in all the dose bins. Denoising substantially improved the DVHs in the direction of the DVHs corresponding to the “clean” MC simulation. This improvement can be more clearly observed from the DVHs from the body (green lines) and PTV (purple lines) regions, since their DVHs are the greatest distance from the antidiagonal. The proposed method demonstrated slightly better results than the UNet-based denoiser.

\begin{figure}
    \centering
    \includegraphics[width=0.8\textwidth]{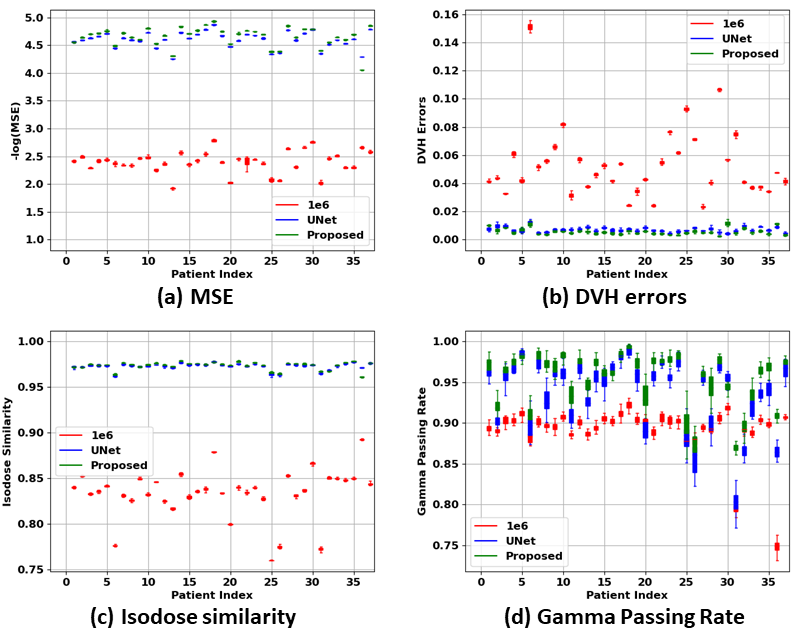}
    \caption{Quantitative comparison in terms of (a) MSE, (b) DVH errors, (c) isodose similarities, and (d) gamma passing rate. The MSE was negatively logarithmically transformed for better visualization. The x-axis is the patient index. The y-axis is the calculated metric value. The boxplot of each data point was calculated based on 15 independent runs of the $1\times{10}^6$ photons-based MC simulation. Different colors indicate the $1\times{10}^6$ photons-based MC simulation (red), the UNet-based denoiser (blue) and the proposed denoiser (green).}
    \label{fig:mse_dvh_isodose}
\end{figure}

Figure~\ref{fig:mse_dvh_isodose} shows patient-specific quantitative comparisons in terms of MSE, DVH error, isodose similarity, and gamma passing rate. For better visualization, we applied a negative logarithmic transformation to the MSE values, as shown in Figure~\ref{fig:mse_dvh_isodose}(a). We can see that both denoisers reduce the MSE by more than two orders of magnitude, but the proposed denoiser is consistently slightly better than the UNet-based denoiser. The short extension of whiskers on the boxplots indicates that both denoisers remove the noise robustly. From the DVH error vs. patient index subplots in Figure~\ref{fig:mse_dvh_isodose}(b), we can observe similar phenomena to the MSE comparisons in Figure~\ref{fig:mse_dvh_isodose}(a): both denoisers substantially reduce the errors due to the contaminated quantum noise, but the proposed denoiser shows better performance than the UNet-based denoiser. Figure~\ref{fig:mse_dvh_isodose}(c) depicts the isodose similarity comparisons. With regard to this metric, the two denoisers show a smaller difference in performance than with the metrics depicted in Figures~\ref{fig:mse_dvh_isodose}(a) and~\ref{fig:mse_dvh_isodose}(b), and both denoisers exhibit obvious better performance than the 15 MC simulations for all patients. Figure~\ref{fig:mse_dvh_isodose}(d) demonstrates the gamma passing rate comparison. One can find that both denoisers show improved results, while the proposed denoiser delivers consistently higher gamma passing rate than the UNet-based denoiser in almost all the evaluated patient cases.

\begin{figure}
    \centering
    \includegraphics[width=\textwidth]{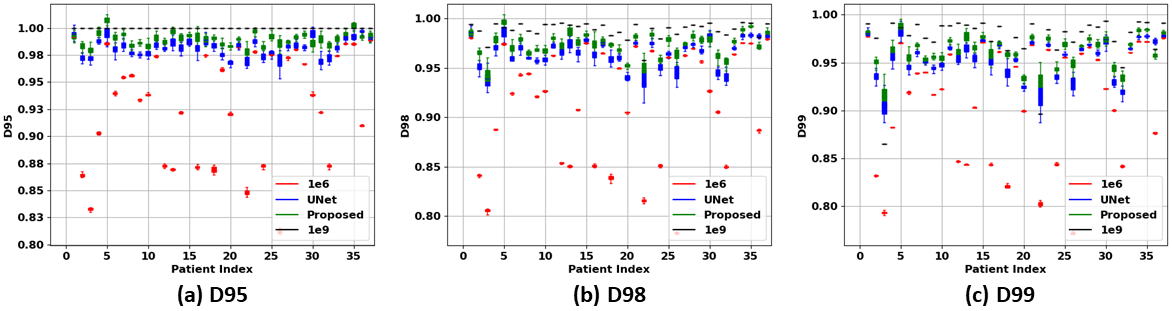}
    \caption{Quantitative comparison in terms of (a) D95, (b) D98, and (c) D99 values. The x-axis is the patient index. The y-axis is the calculated metric value. The boxplot of each data point was calculated based on fifteen independent runs of the $1\times{10}^6$ photons-based MC simulation. Different colors indicate the $1\times{10}^6$ photons-based MC simulation (red), the UNet-based denoiser (blue), the proposed denoiser (green) and the $1\times{10}^9$ photons-based MC simulation (black).}
    \label{fig:d_number}
\end{figure}

To compare the two denoisers in a more clinically relevant way, we present the patient-specific D95, D98 and D99 statistical values among different dose maps in Figure~\ref{fig:d_number}. Because all the dose maps are normalized to the D95 values of the dose maps with respect to the $1\times{10}^9$ photons-based MC simulation, the D95 values associated with the “clean” MC simulation are 1, as shown in Figure~\ref{fig:d_number}(a). As indicated in Figure~\ref{fig:dvh}, the PTV histograms of the noisy dose maps are generally more likely to be antidiagonal; accordingly, the D values associated with the noisy MC simulation are lower than those associated with the “clean” MC simulation. From Figure~\ref{fig:d_number}, we can see that both denoisers yield substantially higher D values. Moreover, compared to the UNet-based denoiser, the D values from the proposed denoiser are consistently closer to those from the $1\times{10}^9$ photons-based MC simulation. Additionally, the proposed denoiser is also more robust than the UNet-based denoiser in terms of having more compact boxplots. 

\begin{table*}[]
    \centering
    \resizebox{\textwidth}{!}{
    \begin{tabular}{ccccccccccc}
    \hline \hline
    & MSE	& DVH  Squared Errors	& D95 &	D98 &	D99 &	Isodose similarity &	Gamma & Time &	FLOPs &	Params \\
    \hline
1e6 &	5.2e-3(7.7e-4) &	5.3e-2(1.1e-4)	& 6.8\% (0.017\%) &	7.3\% (0.012\%) &	6.7\% (0.014\%)	& 0.834(2.50e-4) & 0.892(1.17e-3) &	117ms & 	- &	- \\
UNet &	2.7e-5(1.0e-7)	& 7.2e-3(1.2e-4) &	1.8\% (0.078\%) &	2.4\% (0.098\%) &	2.6\% (0.062\%) &	\textbf{0.973(9.11e-5)} & 0.937(1.65e-3) &	117ms + 454ms &	926G &	49M \\
Proposed	& \textbf{2.5e-5(6.4e-8)} &	\textbf{5.8e-3(1.8e-4)} &	\textbf{1.1\% (0.067\%)} &	\textbf{1.6\% (0.089\%)} &	\textbf{1.9\% (0.068\%)} &	\textbf{0.973(9.29e-5)} & \textbf{0.954(1.55e-3)}	&\textbf{117ms + 39ms} &	\textbf{55G}	& \textbf{12M} \\
1e9	 &-	 & -	& - & 	-	& -	& - & -	& 120s &	-	& - \\
\hline \hline
    \end{tabular}}
    \caption{Quantitative comparisons among different dose maps against the benchmark dose maps associated with the $1\times{10}^9$ photons-based MC simulation. The values are calculated as the mean values and standard deviations (inside the parentheses) across 15 independent MC simulations. The time are calculated as the MC simulation time plus the denoising time, where applicable. }
    \label{tab:metrics}
\end{table*}

Table~\ref{tab:metrics} presents different metrics quantifying the performance for comparison. Not surprisingly, both denoisers improved the performance substantially over the noisy dose maps. More specifically, with regard to the mean performance, the proposed denoiser produced numerically better results than the UNet-based denoiser in each of the metrics, except the isodose similarity metric, for which both denoisers showed the same performance. As for the robustness, both denoisers showed comparable performance, considering that each denoiser performed better on three of the six metrics. It should be noted that, for each metric, the standard deviation is considerably lower than the associated mean value, which suggests that both denoisers are very robust.

With regard to efficiency, the proposed denoiser substantially outperformed the UNet-based denoiser, as it exhibited an $\sim$11.6 times increase in inference speed (39 ms vs. 454 ms). To compare the theoretical acceleration factors, Table~\ref{tab:metrics} lists the FLOPs of both denoisers: the proposed denoiser has ~16.8 times fewer float operations per forward propagation than the UNet-based denoiser (55G vs. 926G). The proposed architecture also has ~4 times fewer parameters than the UNet-based denoiser, which further explains its greater efficiency. Thus, despite the proposed deep dose plugin’s lighter architecture, it performs better than the conventional UNet-based denoiser in terms of the six different quantitative metrics used in this work (Table~\ref{tab:metrics}).

\section{Discussion and Conclusions}
Monte Carlo computation time can be reduced from the order of hours to the order of minutes by adopting the advanced GPU parallel computation framework and various algorithm acceleration strategies\citep{RN23}. This time can be further decreased to the order of seconds if one first conducts a fast and noisy MC simulation and then denoises the resulting dose map with a DL-based denoiser\citep{RN48}. The goal of this work is to move one step further by improving MC dose calculation efficiency to a real-time level (sub-seconds), which would be of significance to clinical applications such as IMRT iterative plan optimization and online adaptive re-planning. We have proposed a DL-based MC dose denoiser that can process an input dose volume with a size of $256\times256\times64$ in 39 ms, as shown in Table~\ref{tab:metrics}. This simple denoiser has plug-and-play functionality, so it can be attached as a plugin in current MC dose engines without interrupting the whole pipeline; hence, we have termed it “deep dose plugin.” We would like to highlight that the whole pipeline takes around ~0.15 s, even taking the noisy MC simulation time (117 ms) into consideration. This time competes with any of the fastest dose engines, including GPU-based pencil beam–based dose engines\citep{RN60}. Although we only used the prostate cancer patients  dataset as a showcase, the proposed method is not dependent on tumor sites and can be readily applied to any other patient cases. As such, we expect that an MC dose engine featuring our deep dose plugin could be routinely used in a modern TPS for fast yet accurate dose calculations.

Two key modifications are responsible for achieving this real-time denoising efficiency. One is the adoption of the voxel shuffle/unshuffle operators. We were motivated to use these two operators by our observation that the major computational burden stems from the input and the output layers, where the associated features have the largest sizes. One may use a smaller volume size as the input to reduce the computational burden, but this would inevitably lead to inferior performance due to the information loss. As for the output layer, in order to save computational resources, many networks predict a smaller output, which is then upsampled to the original output size. This strategy is very popular in fields like natural image segmentation \citep{RN46} and human pose estimation \citep{RN63}, where the performance is not very sensitive to the output size. However, for our dose denoising task, the performance would be degraded during the upsampling process. By contrast, the voxel shuffle/unshuffle operators used in this work can decrease the input and output size substantially, thereby accelerating the process, while avoiding performance degradation by retaining all the information as well as the receptive field from the original input. Indeed, the voxel shuffle idea was also used in the natural image super resolution field to improve the model’s inference efficiency \citep{RN68}. In that work, the authors used a pixel shuffle layer in the output side to accelerate the model. By contrast, in our work, we simultaneously attached a voxel unshuffle layer and a voxel shuffle layer to the input layer and the output layer, respectively, to achieve further acceleration. Another effect of using these operators to preprocess the input and postprocess the output is that they use fewer downsampling and upsampling modules to reach same receptive field size, because the resultant size is only half of the original size in each dimension. This can further reduce the computational burden to some extent. We believe this simple acceleration strategy can be easily extended to other architectures and other applications.

Another important strategy for accelerating the model is decoupling the regular 3D volumetric convolution operator into a 2D axial convolution and a 1D slice convolution. Basically, any given 3D convolution operator $W_r\in R^{\mathrm{C_{out}}\times \mathrm{C_{in}}\times K_h\times K_w\times K_d}$ will try to characterize the correlations spanning in 4 different directions: channel, height, width, and slice. As such, one can change the convolution method in each direction to accelerate the process. One straightforward way is to decouple the 3D convolution operator into 4 different 1D convolution operators to characterize the correlations separately and consecutively in each direction, just like decoupling a 2D Gaussian kernel into two 1D Gaussian kernels for the purpose of acceleration. Theoretically, this way can achieve the greatest acceleration factor via decoupling. However, we have found that the efficiency of GPU utilization is very low when decoupling in this straightforward way. To be specific, in the channel direction, the computation is indeed not a real convolution operation. In detail, when the size $C_{in}$ of the 3D convolution operator in the channel direction is equal to the number of input channels, its computation equals a special convolution case where the kernel size is same as the size of the signal. However, when $C_{in}$ is not equal to the number of input channels, it should be divisible by the numbers of both the input channels and the output channels to be supported by current mainstream training frameworks such as PyTorch \citep{paszke2019pytorch}, and is termed a group convolution. Group convolution will first divide both the input and the output into several groups, then perform the regular convolution for each group. The limit of group convolution is when the group number is equal to the number of input channels, i.e., the so-called depth-wise convolution, which is widely used in lightweight architectures, such as MobileNet \citep{RN70} and ShuffleNet \citep{RN71}, designed for mobile devices. However, we found that the inference speed was much lower than with regular 3D convolution when we used the depth-wise convolution in our network. This is because the GPU’s computational efficiency of depth-wise 3D convolution is not well-optimized for current deep learning frameworks. Therefore, to be compatible with current mainstream deep learning frameworks, we kept the regular computation method unchanged in the channel direction in this work. With regard to the convolution method in the other three directions, one can directly use three 1D convolutions to simulate the regular 3D convolution to characterize their correlations. However, we have found that these also slightly degrade the GPU’s computational efficiency, which might be due to the over-fragmented computation, which is not friendly to GPUs. Therefore, in this paper, we used one 2D convolution to characterize the axial correlations, followed by one 1D convolution to characterize the depth correlation. One advantage of decoupling the regular 3D convolution into one 2D and one 1D convolution is that it doubles the number of nonlinear layers and consequently improves the nonlinear expressiveness. This might partially explain why the proposed denoiser performs better than the conventional UNet-based denoiser despite being lighter and faster.

The above acceleration strategies are orthogonal to other strategies, such as the low-bit–based quantization technique \citep{RN72} and the use of advanced inference engines such as TensorRT \citep{migacz2017nvidia}. Therefore, they can be combined to achieve further acceleration. However, it is worth mentioning that, by using the proposed acceleration strategies, we have achieved real-time efficiency (sub-seconds) for the whole dose calculation pipeline: 117 ms MC simulation time plus 39 ms denoising time. Therefore, one may argue that there is no reason for further acceleration for this particular clinical task.
The efficiency related data listed in Table~\ref{tab:metrics} were obtained by using Python script, and they only consider the pure MC simulation time and the denoising time. Therefore, the denoiser will likely be faster if more efficient programming languages are used. In practice, as a whole pipeline, we should also consider potential data exchange time among the hard-drive, RAM memory and GPU memory. Since the noisy dose map should already be in the GPU memory once the GPU-based MC simulation is finished, this data exchange time for the denoising step should be negligible.

Regarding the performance, the qualitative comparisons depicted in Figures~\ref{fig:diff_profile}-\ref{fig:dvh} show that both denoisers can suppress the noise effectively and produce dose maps comparable to the benchmark dose maps. From the quantitative comparisons presented by the patient-specific metrics in Figures~\ref{fig:mse_dvh_isodose} and~\ref{fig:d_number}, as well as the overall metrics in Table~\ref{tab:metrics}, one can further verify that the proposed denoiser numerically exhibits slightly, but consistently, better performance than the conventional UNet-based architecture, though this performance difference may not be clinically significant.

Lastly, we want to emphasize that both denoisers were trained in a noise-to-noise weakly supervised learning framework, where both the input and the output were noisy dose maps from $1\times{10}^6$ photons-based MC simulations. One advantage of this method is that it requires substantially less data generation time than the conventional noise-to-clean supervised learning framework. As suggested in Table~\ref{tab:metrics}, for each data point, one only needs 117 ms to generate the noisy dose map. This feature might be important when deploying the trained models in clinical practice through fast fine-tuning–based adaptation. When combined with the proposed lightweight architecture, for which one complete forward and backward propagation takes ~0.1 s, this fine-tuning process can be finished in $<$20 min given 100 patients for data generation and $1\times{10}^4$ iterations for fine-tuning. Another advantage of the noise-to-noise weakly supervised learning framework is that it allows us to randomly exchange the input noisy data and the target noisy data for model training and can, thus, double the training dataset size.

In summary, in this paper, we developed a deep learning–based fast MC dose denoiser that can be trained with a weakly supervised learning framework. This denoiser can be plugged into current GPU-based MC dose engines and can finish the whole dose calculation pipeline in ~0.15 s. To achieve this goal, we first applied the voxel shuffle and unshuffle layers to reduce the input and output sizes without compromising the amount of information and second, decoupled the regular 3D volumetric convolution into one 2D axial convolution and one 1D slice convolution to reduce the computation complexity.
\section*{Data Access}
The data analysed during the current study are not publicly available for ethical reasons but are available from the corresponding author on reasonable request and approved by the IRB.

\section*{Acknowledgment}
We would like to thank Varian Medical Systems Inc. for supporting this study and Dr. Jonathan Feinberg for editing the manuscript. The dataset used in this paper were collected in UT Southwestern Medical Center, and approved by the IRB.

\appendix
\section{Equivalence of the noise-to-clean-based cost functions and the noise-to-noise-based cost functions}
\label{appendix: equivalence}
From cost function~(\ref{eq:DL_objective}), we have the following equivalent cost function:

\begin{tiny}
\begin{eqnarray}
     \bar{W} &= &\arg{\min_W{\int_{X}\int_{\epsilon_1}\int_{\epsilon_2}\left|\left|\Phi_W\left(X+\epsilon_1\right)-\left(X+\epsilon_2\right)+\epsilon_2\right|\right|_2^2d_{\epsilon_2}d_{\epsilon_1}dX}} \nonumber \\
             &= &\arg{\min_W{\int_{X}\int_{\epsilon_1}\int_{\epsilon_2}\left\{\left|\left|\Phi_W\left(X+\epsilon_1\right)-\left(X+\epsilon_2\right)||_2^2+2\left[\Phi_W\left(X+\epsilon_1\right)-\left(X+\epsilon_2\right)\right]\epsilon_2+||\epsilon_2\right|\right|_2^2\right\}d_{\epsilon_2}d_{\epsilon_1}dX}}\nonumber\\
             &= &\arg{\min_W{\int_{X}\int_{\epsilon_1}\int_{\epsilon_2}\left\{\left|\left|\Phi_W\left(X+\epsilon_1\right)-\left(X+\epsilon_2\right)\right|\right|_2^2+2\Phi_W\left(X+\epsilon_1\right)\epsilon_2\right\}d_{\epsilon_2}d_{\epsilon_1}dX}}\nonumber\\
             &= &\arg{\min_W{\int_{X}\int_{\epsilon_1}\int_{\epsilon_2}\left|\left|\Phi_W\left(X+\epsilon_1\right)-\left(X+\epsilon_2\right)\right|\right|_2^2d_{\epsilon_2}d_{\epsilon_1}dX+2\int_{X}\int_{\epsilon_1}{\Phi_W\left(X+\epsilon_1\right)}d_{\epsilon_1}dX\int_{\epsilon_2}{\epsilon_2d\epsilon_2\ }}}\nonumber \\
             &= &\arg{\min_W{\int_{X}\int_{\epsilon_1}\int_{\epsilon_2}\left|\left|\Phi_W\left(X+\epsilon_1\right)-\left(X+\epsilon_2\right)\right|\right|_2^2d_{\epsilon_2}d_{\epsilon_1}dX}} 
\label{eq:derivation}
\end{eqnarray}
\end{tiny}

In the process of this proof, the constant values that are independent of the variable $W$ are ignored. In addition, we have assumed that $\Phi_W(X+\epsilon_1)$ is energy bounded, and the expectation of noise $\epsilon_2$ is zero, i.e., $\int_{X}\int_{\epsilon_1}{\Phi_W\left(X+\epsilon_1\right)dXd_{\epsilon_1}}<\ \infty$ and $\int_{\epsilon_2}\epsilon_2d\epsilon_2=0$. This loose condition can be readily satisfied by our MC dose denoising task. From cost functions~(\ref{eq:DL_objective}) and~(\ref{eq:derivation}), one can find that both the conventional noise-to-clean training strategy and the introduced noise-to-noise training strategy have equivalent cost functions, and thereby share same global minimum.

\section{Modified UNet architecture}
\label{appendix: architecture}
As illustrated in Figure~\ref{fig:network}, the proposed architecture consists of four different modules—multichannel input, backbone, feature pyramid, and prediction head.

\textbf{Multichannel Input:} Generally speaking, the larger the input image size, the more layers are required to keep the same receptive field size, and consequently, the heavier the computational burden is. Specifically, we have found that the major computational burden of the 3D UNet architecture is on the first downsampling layer and the last upsampling layer due to the larger input and output sizes. To alleviate this problem, we first equipped our model with a voxel unshuffle layer on the input side, which leads to a multichannel input for further processing. To explain voxel unshuffle in detail, let us assume the original input has a dimension of $B\times C\times H\times W\times D$, where $B$ and $C$ represent the batch size and the channel number, respectively. For our MC dose denoising task, we set $C=1$. After the voxel unshuffle layer, the input is rearranged into an 8-channel input with a dimension of  $B\times8\times\frac{H}{2}\times\frac{W}{2}\times\frac{D}{2}$. In each channel, the sub-volume is a downsampled version of the original input with a step size of 2 and a different starting point, i.e., $X_{i,j,k}=X\left[i::2,j::2,k::2\right],{i,j,k}\in[0,1]$ if using Python programming language, where $X_{i,j,k}$ denotes the $\left(4k+2j+i\right)_{\mathrm{th}}$ channel. A simplified illustration of the 2D version is provided in Figure~\ref{fig:network}. This newly introduced multichannel input still contains all the information, and thus keeps the same receptive field, as the conventional one-channel–based input. However, this multichannel input can improve the model efficiency substantially since the input size is two times smaller in each dimension. Even though the channel number is larger, the associated increase in computational burden is trivial.

\textbf{Backbone:} The backbone might be one of the most important components in the modern CNN architecture, since it has the greatest influence on the feature quality. It can be any of the sophisticated designs that have been investigated, such as VGG \citep{RN52}, Inception \citep{RN53}, ResNet \citep{RN44}, or ResNeXT \citep{RN54}. Basically, it should be deep enough to ensure the network’s nonlinear expressiveness. However, too many layers would hamper the computational efficiency, so a tradeoff must be made between the network depth and the associated computational efficiency. The conventional UNet-based architecture adopts a simple backbone consisting of several downsampling modules that are consecutively stacked for feature extraction. Each module is composed of three successive operators: 1) a convolution (CONV) operator, 2) a normalization operator, and 3) a rectified linear unit (ReLU). The major computational complexity of each layer stems from the convolution operator. For our MC dose denoising task, a 3D volumetric convolution operator should be used to characterize the correlation spanning three different dimensions, which makes the problem of computational burden more pronounced than in conventional 2D scenarios. To overcome this challenge, in this work, we decoupled the regular 3D volumetric convolution operator into a 2D axial convolution operator and a 1D slice convolution operator. Let us represent the regular convolution operator with the same kernel size $K$ in each dimension as $W_r\in R^{C_{in}\times C_{out}\times K\times K\times K}$, where $C_{in}$ and $C_{out}$ denote the input and output channels, respectively. This is used to characterize the axial correlations and the slice correlations simultaneously. By contrast, our decoupled convolution operator is designed to simulate the regular 3D convolution by first using a 2D convolution operator $W_a\in R^{C_{in}\times C_{out}\times K\times K\times1}$ to describe the axial spatial correlations, then using a 1D convolution operator $W_d\in R^{C_{out}\times C_{out}\times1\times1\times K}$ to describe the slice correlations. This decoupling process is equivalent to decomposing a high-dimension convolution into two low-dimension convolutions, which accelerates the process. Therefore, in our architecture, each module is composed of six operators. The first three operators are axial CONV-normalization-ReLU, and the second three operators are slice CONV-normalization-ReLU.

It should be mentioned that in the conventional UNet-based architecture, the stride is 2 in each dimension of the volumetric convolution operator. In our architecture, given the same input feature size, making the output feature size the same as the regular volumetric convolution operator-based module requires that the strides are 2 in the axial dimensions and 1 in the slice dimension for the axial convolution operator, and the opposite for the slice convolution operator. Given an input size of $256\times256\times64$, which we used in this work, the conventional UNet-based architecture needs six downsampling modules to reach the largest receptive field, since the smallest dimension is 64; however, the proposed architecture only needs five downsampling modules to reach the same receptive field because the smallest dimension is 32, since it uses the multichannel input. Using fewer downsampling modules can further reduce the computational burden. However, fewer downsampling modules do not suggest weaker nonlinear expressiveness. By contrast, our backbone contains ten nonlinear layers, since each module consists of two nonlinear layers, while the backbone of the conventional UNet contains only six nonlinear layers.

\textbf{Feature Pyramid:} To fully exploit the features that contain information of different scales, ranging from rich semantic information (top features) to fine detailed information (bottom features), our architecture uses a feature pyramid module for feature fusion, similar to the UNet-based architecture. As shown by the gray shaded module in Figure~\ref{fig:network}, we first use the computation-efficient bilinear operation to upsample the low-resolution features. These are then further preprocessed by two convolution operators before being concatenated with the features from the backbone part that have same resolutions. Again, in this feature pyramid module, the axial and slice convolution operators are applied to capture the spatial correlations spanning the axial direction and the slice direction, respectively.

\textbf{Prediction Head:} As mentioned above, the output feature has the largest size of all the layers. Therefore, the computational complexity is also considerably higher than the other layers with smaller feature sizes. To accelerate the prediction head, we first use the stacked 2D convolution operator and 1D convolution operator to generate an 8-channel output, then we use a 3D voxel shuffle operator to restore the final 1-channel dose map output, which is the inverse operator of the voxel unshuffle operator detailed above.

In addition, for the normalization layer, we always use the instance normalization \citep{RN55} as the normalization operator in this work. This can also be replaced by other normalization operators, such as batch normalization \citep{RN56} and group normalization \citep{wu2018group}.

For the first module in the backbone, the feature numbers are set to 64 for both the axial and slice convolution operators. For the modules that follow in the backbone, the feature numbers will be doubled, as the feature sizes are decreased by half. As for the modules in the feature pyramid, we follow the same feature number selection strategy as the conventional feature-concatenation–based UNet architecture. All the axial convolution operators have a kernel size of $3\times3\times1$ with strides of 2 in the first two dimensions and a stride of 1 in the last dimension. All the slice convolution operators have a kernel size of $1\times1\times3$ with strides of 1 in the first two dimensions and a stride of 2 in the last dimension. 

For a better description about the network, we further tabulate the configuration parameters as well as the associated FLOPS for all the convolutional operators in Table~\ref{tab:layer_configuration}. From this table, we can find that the efficiency boost of the proposed network over the original UNet is attributed to both the introduced shuffle/unshuffle operators and the 3D convolution decomposition strategy. The former can reduce the number of the channels given same feature map size as the original UNet (or reduce the feature map size given same number of channels), while the latter can reduce the convolutional kernel size.

\begin{table}[]
    \centering
    \begin{tabular}{c}
        \includegraphics[width=0.9\textwidth]{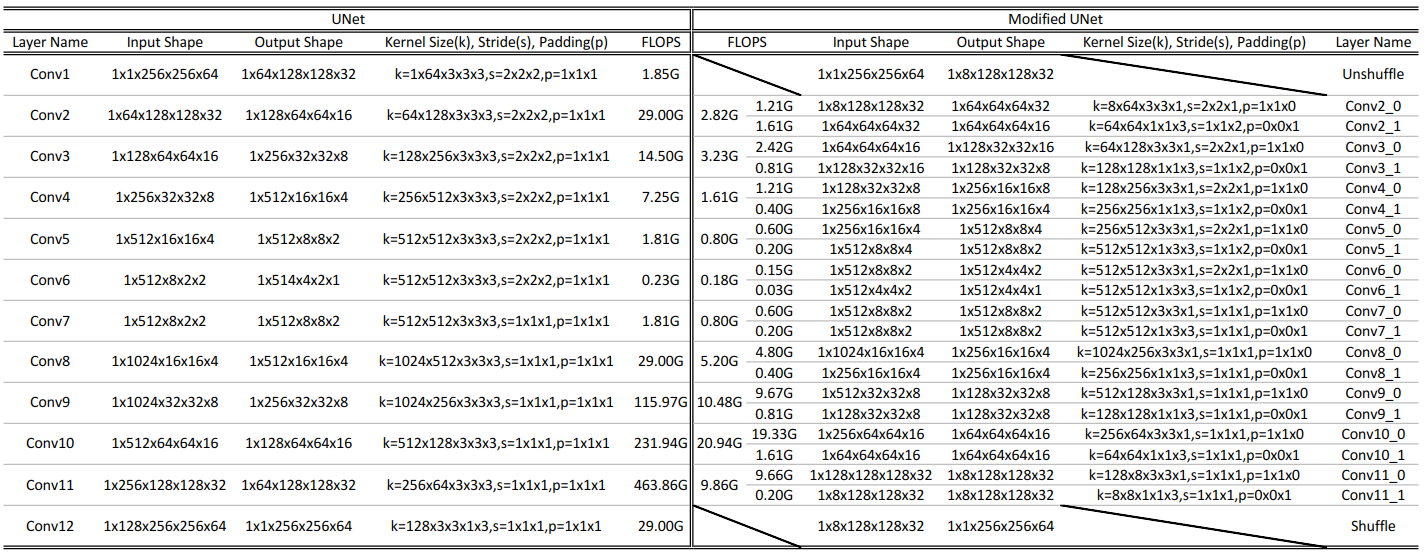} 
    \end{tabular}
    \caption{Layer configuration parameters as well as the number of the float operations (FLOPS) for the UNet and the modifed UNet. The input and output shapes are organized as $\mathrm{Batch\ Size}\times \mathrm{Channel} \times \mathrm{Height} \times \mathrm{Width} \times \mathrm{Depth}$. The first two numbers of the kernel size denote the input and output channels. The last three numbers of the kernel size denote the real convolutional kernel size. The stride and the padding sizes represent the sizes in three different dimensions. The FLOPS is in the unit of $G=10^9$.}
    \label{tab:layer_configuration}
\end{table}

\bibliography{./deepdoseplugin}
\bibliographystyle{agsm}
\end{document}